# Towards a conceptual framework for innate immunity

Jamie Twycross, Uwe Aickelin

**Abstract.** Innate immunity now occupies a central role in immunology. However, artificial immune system models have largely been inspired by adaptive not innate immunity. This paper reviews the biological principles and properties of innate immunity and, adopting a conceptual framework, asks how these can be incorporated into artificial models. The aim is to outline a meta-framework for models of innate immunity.

## 1 Introduction

Immunology has traditionally divided the immune system into innate and adaptive components with distinct functional roles. For many years, research was focused on the adaptive component. However, the prevailing view in immunology now shows the innate system to be of central importance [1]. The first part of this paper focuses on the innate immune system and on ways in which it interacts with and controls the adaptive immune system and discusses research over the last decade which has uncovered the molecular basis for many of these mechanisms, reviewed in [2]. It first contrasts the innate and adaptive immune systems and briefly reviews essential biology. It then discusses specific mechanisms of interaction between cells of the innate and adaptive immune systems, and concludes by showing how these mechanisms are examples of more general systemic properties.

While the integral role of the innate immune system has been established in immunology, artificial immune system models, surveyed in [3, 4], have largely taken their inspiration from adaptive immunity. The second part of this paper adopts the conceptual framework of Stepney et al. [5] and addresses how ideas from innate immunity might be modelled in artificial immune systems. The conceptual framework is first briefly summarised and then a general meta-framework for models incorporating innate immunity is presented and refined through the discussion of specific models properties.

## 2 Innate immunity

This section begins with an overview of well-established conceptions of innate immunity. Research which over the last decade has served to highlight the central role of the innate immune system is then discussed. Lastly, general properties of the innate immune system which have been drawn out by this research are presented. Review papers as well as the original articles are cited, and original figures are reproduced to enhance the necessarily brief summaries of the mechanisms.

## 2.1 Contrasting innate and adaptive immunity

Differences between the innate and adaptive immune systems can be seen on a number of levels (Table 1). The adaptive immune system is organised around two classes of cells: T cells and B cells, while the cells of the innate immune system are much more numerous, including natural killer (NK) cells, dendritic cells (DCs), and macrophages. The receptors of innate system cells are entirely germline-encoded, in other words their structure is determined by the genome of the cell and has a fixed, genetically-determined specificity. Adaptive immune system cells possess somatically generated variable-region receptors such as the TCR and BCR (T and B cell receptors) with varying specificities, created by a complex process of gene segment rearrangement within the cell. On a population level, this leads to a non-clonal distribution of receptors on innate immune system cells, meaning that all cells of the same type have receptors with identical specificities. Receptors on adaptive immune system cells however, are distributed clonally in that there are subpopulations of a specific cell type (clones) which all possess receptors with identical specificities, but that generally, cells of the same type have receptors with different specificties [1, 6, 7].

| property | innate immune system | adaptive immune system |
|---|---|---|
| cells | DC, NK, macrophage. | T cell, B cell. |
| receptors | germline-encoded. rearrangement not necessary. non-clonal distribution. | encoded in gene segments. somatic rearrangement necessary. clonal distribution. |
| recognition | conserved molecular patterns. selected over evolutionary time. | details of molecular structure. selected over lifetime of individual. |
| response | cytokines, chemokines. | clonal expansion, cytokines. |
| action time | immediate effector activation. | delayed effector activation. |
| evolution | vertebrates and invertebrates. | only vertebrates. |

Table 1: Differences between innate and adaptive immunity.

The molecules which a receptor is able to bind with and recognise are known as ligands. While all receptors at the most basic level recognise molecules, ligands are often discussed in terms of higher-level structures. The variable-region receptors of adaptive immunity recognise features of pathogen structure, with BCRs directly recognising peptide sequences on pathogens, such as components of bacterial cell membranes, and TCRs recognising peptide sequences which have first been processed by DCs. These receptors are selected for over the lifetime of the organism by processes such as clonal expansion, deletion or anergy and are under adaptive not evolutionary pressure. Conversely, innate immune system receptors recognise a genetically-determined set of ligands under evolutionary pressure. One key group of innate receptors is the pattern recognition receptor (PRR) superfamily which recognises evolutionary-conserved pathogen-associate molecular patterns (PAMPs). PRRs do not recognise a specific feature of a specific pathogen as variable-region receptors do, but instead recognise common features or products of an entire class of pathogens. The immune system utilises adaptation of variable-region receptors to keep pace with evolutionary more rapid pathogens [1, 6].

The environment of a cell in vivo is the tissue in which it is located. Tissue is formed by specialised groups of differentiated cells, and itself forms major components of organs. A substantial part of tissue volume is extracellular space and filled by a structured network of macromolecules called the extracellular matrix. Many of the molecules found in the extracellular matrix are actively produced by cells and involved in intercellular signalling [8, 9]. Cytokines are secreted molecules which mediate and regulate cell behaviour, two important subsets of which are tissue factors, inflammation-associated molecules expressed by tissue cells in response to pathogen invasion, and chemokines, cytokines which stimulate cell movement and activation. Cytokines bind to germline-encoded cytokine receptors present on all cells and are widely produced and consumed by both innate and adaptive immune system cells during an immune response. Recognition by the innate immune system leads to the immediate initiation of complex networks of cytokine signalling which orchestrate the ensuing immune response. Adaptive responses additionally involve processes of cell selection such as clonal expansion, deletion and anergy, which take several days [1, 6].

## 2.2 Recent developments

This section reviews key developments over the last decade in our understanding of the innate immune system. Over this period, intense research has highlighted the central role of the innate system in host defense through its interaction with the adaptive immune system and with tissue, and uncovered the molecular basis for these interactions. These developments have lead immunologists to reevaluate the roles of both the innate and adaptive immune systems in the generation of immunity, installing innate immunity as a vital component in the initiation and modulation of the adaptive immune response [2].

NK cells of the innate immune system respond to the disruption of normal cell physiology in what has been termed the "missing self" model of NK cell activation [10]. Most normal tissue cells constitutively express MHC class I molecules, which present intracellular host-derived peptides on the cell surface. Presentation of virus-derived peptides leads to activation of CTL (cytotoxic T lymphocyte) cells and apoptosis in the infected cell through ligation with the TCR of the CTL [11]. However, viruses and other infectious agents have evolved to interfer with MHC class I antigen presentation [12] and so evade a CTL response. In the "missing self" model (Figure 1), NK cells are activated either by reduced signalling through receptors of the KIR family, inhibitory receptors specific for host MHC class I, leading to apoptosis of the cognate cell [13]. This creates a no-win situation for the virus: if MHC class I expression is unaffected, it will be open to detection and removal through a CTL-based adaptive immune response, but if it affects MHC class I expression, it will be open to detection and removal through an NK-based innate immune response.

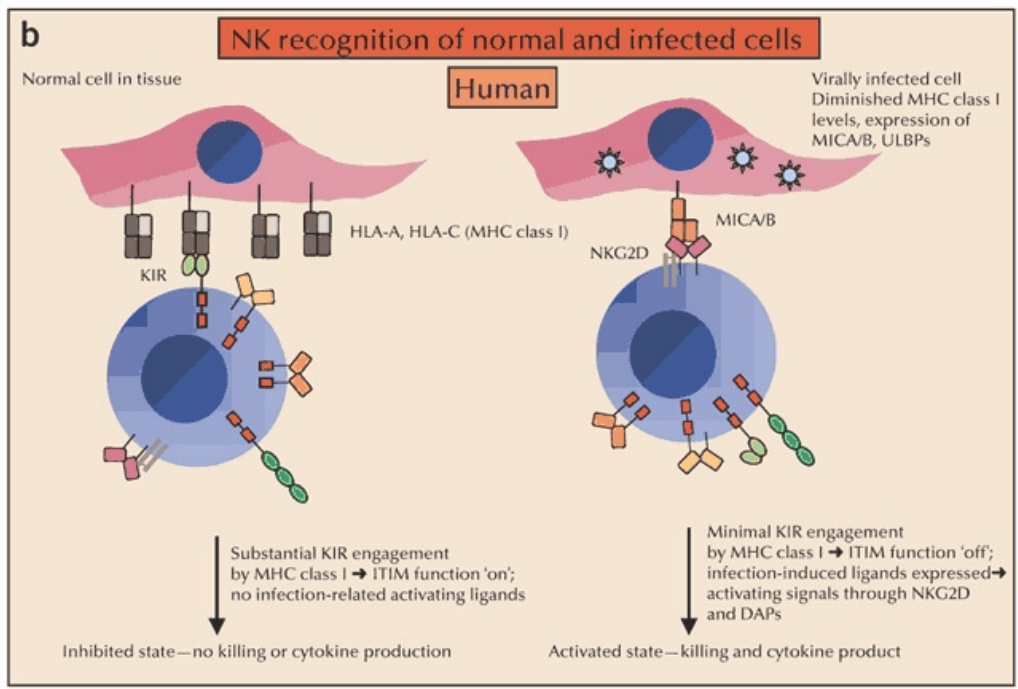

Fig. 1: NK receptors and NK recognition, from [2]

Some of the most exciting recent advances have been made in uncovering the role of TLRs in determining DC differentiation and so a mechanism by which the innate immune system mediates the quality of an adaptive immune system response [2, 15] (Figure 2). Initial ligation by different PAMPs and tissue factors of different TLRs on DCs "primes" DCs to differentiate along different pathways, resulting in mature and immature DCs which produce different Th (T helper) cell polarisation factors. Release of these polarisation factors upon interaction with naive T cells causes the naive cell to differentiate into Th1, Th2 or Treg cells, all distinct types of T cell [14]. DCs, through TLRs, couple the quality of the adaptive immune effector response to the nature of the pathogen. Other PRR receptor families have also been implicated in Th polarisation [16]. Interestingly, recent research [17] suggests a renewed role for variable-region receptors not just in the determination of the antigen specificity of an immune response, but also in the regulation of this response. In place of the purely "instructive" DC to T cell paradigm, the responding Th1 or Th2 cells reinforce signals to B cell or CTL effectors in a "success-driven" consensual model of T cell polarisation.

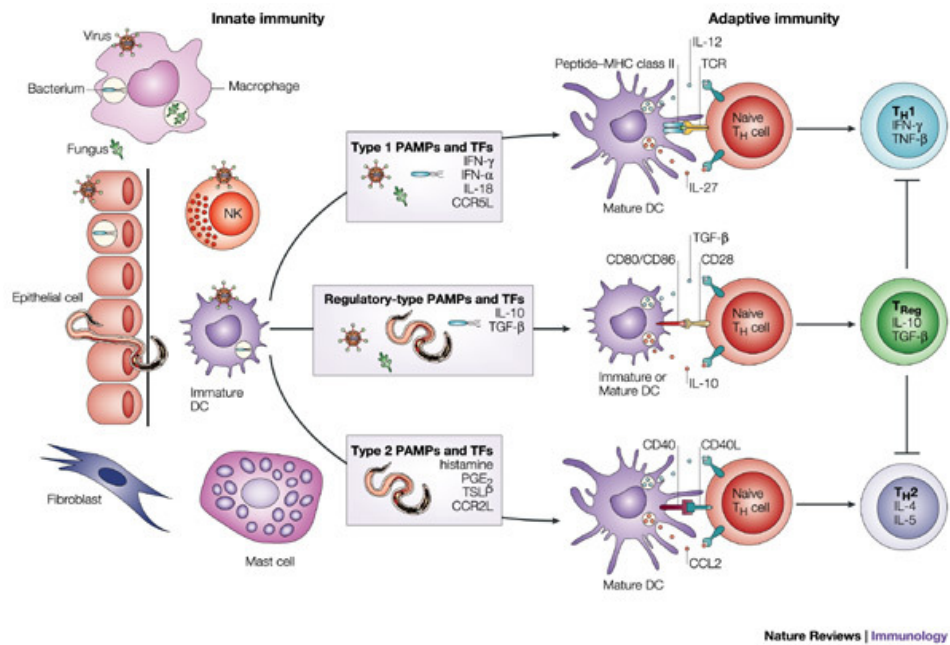

Fig. 2: DC polarisation of Th cells, from [14]

As well as polarising Th cells, DCs play a key role in maintenance of populations of T cells. Tolerance is the ability of the immune system to react in a non-biodestructive manner to stimuli and has long been associated with adaptive immunity. Tolerance is usually discussed in terms of apoptosis or anergy of self-reactive T and B cells, and was initially proposed to occur centrally in a relatively short perinatal period, as epitomised in the clonal selection theory of Burnet [19, 20]. While recent research shows the continuing importance of central tolerance mechanism [21], it is now accepted that peripheral tolerance mechansims which operate to censor cells throughout the lifetime of the host are of equal importance. DCs of the innate immune system lie at the heart of the generation of peripheral tolerance. Models propose that DCs continually uptake apoptotic and other material from peripheral tissues under normal steady-state, nonpathogenic conditions. Periodically, DCs migrate to draining lymph nodes where they delete lymphocytes by presenting the processed material which, as representative of tissue in the absence of pathogen, needs to be tolerated by the host [22]. Signals received by DCs 'license' [18] (Figure 3) them to promote either T cell clonal expansion, or T cell clonal anergy or deletion. Research has established the molecular basis for such models. The absence of TLR signalling on DCs [22, 23] or the presence of signalling through receptors involved in the up-take of apoptotic material [24, 25] leads to distinct semimature and mature DC populations which interact with T cells to promote tolerance or immunogenicity respectively.

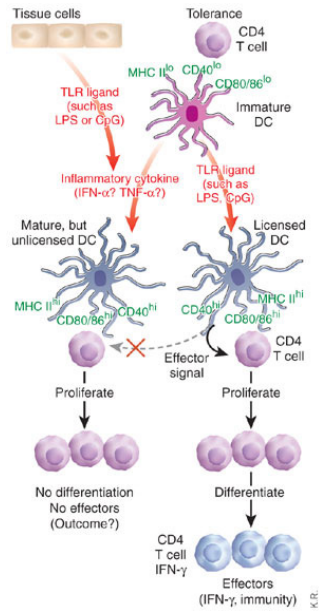

Fig. 3: DC Th tolerance, from [18]

Cosignalling receptors and their ligands provide another mechanism by which DCs determine the qualitative and quantitative nature of adaptive immune responses. CD80 and CD86 are costimulatory molecules expressed on DCs and bind with the CD28 and CTLA-4 cosignalling receptors on Th cells. Binding to CD28 leads to upregulation of Th activity and an immunogenic response, whereas CTLA-4 binding to downregulation of activity and tolerance. CD28 is constitutively expressed by Th cells, whereas the latter in proportion to the strength of TCR stimulation. CD80 and CD86 do not bind equivalently to CD28 and CTLA-4, and through selective expression by DCs of these molecules, innate immune system cells initiate and regulate Th cell activity. A key concept which has emerged from this research is the importance of sequential and properly timed interactions in the development of an immune response [2, 26, 27].

## 2.3 Summary

As the biology described in this section shows, the protection afforded to the host by the immune system as a whole arises from mechanisms of the innate and adaptive immune systems, which help form an integrated system of host protection. While there can be no doubt that specific recognition by the adaptive immune system plays an important role in functions such as pathogen recognition and removal, it is now clear that innate immune system mechanisms play an equally important role. The mechanisms discussed above are specific examples of more general properties of innate and adaptive immune system function and interaction, which are summarised in Table 2.

| | |
|---|---|
| property 1 | pathogens are recognised in different ways by the innate and adaptive immune systems. |
| property 2 | innate immune system receptors are determined by evolutionary pressure. |
| property 3 | response to pathogens is performed by both the innate and adaptive systems. |
| property 4 | the innate immune system initiates and directs the response of the adaptive immune system. |
| property 5 | the innate immune system maintains populations of adaptive immune system cells. |
| property 6 | information from tissue is processed by the innate immune system and passed on to the adaptive immune system. |

Table 2: General properties of the innate immune system.

Considering the innate as well as adaptive immune system highlights how immune system cells interact with pathogens on multiple levels (Property 1). While the variable-region receptors of adaptive immunity are often specific for one feature of one particular pathogen, germline-encoded receptors such as PRRs of innate immunity are specific for features belonging to an entire class of pathogens. Innate immune system cells also respond not only to pathogen structure, but also to pathogen behaviour, either directly through PAMPs and TLRs, or indirectly through changes in tissue cell behaviour (NK cells). Innate receptor specificity is determined by evolutionary pressures, whereas adaptive processes such as peripheral tolerance determine the range of specificities of adaptive receptors (Property 2).

Innate immune system cells, as well as recognising pathogen, respond to them directly (Property 3), as with NK cell monitoring of MHC class I expression. Such recognition and response mechanisms when taken together show how the innate and adaptive immune systems work together to provide a broad coverage of protection to the host. Recognition by the innate immune system does not usually lead to a solely innate response, but instead also initiates and modulates an adaptive response through DC polarisation of Th cells and modulation of costimulatory signals (Property 4). Mechanisms such as DC tolerisation of Th cells, as well as relying on antigen processed by DCs, also shows how innate immune system cells maintain populations of cells (Property 5). The adaptive response is driven by information not only directly sensed by adaptive immune system cells, but equally by information gathered and processed by innate immune system cells, as with DC collection, processing and presentation of antigen to T cells (Property 6).

## 3 Modelling innate immunity

As artificial immune systems develop in their sophistication and so are more able to realise the functions of biological immune systems, they will need to incorporate properties of innate immunity into their models. This section first reviews the conceptual framework for artificial immune systems of Stepney et al. [5]. Adopting this framework and drawing on the biology of the previous section, it then proposes a number of general properties of models incorporating innate immunity. Looking first at the mechanisms of the previous section as a whole, and then individually, these general properties are discussed and refined. The aim is to suggest a meta-framework which highlights the key properties of models in general and how they might be realised in various individual models.

### 3.1 Conceptual frameworks

In [5], Stepney et al. present a conceptual framework within which biologically-inspired models and algorithms can be developed and analysed. Figure 4 summarises their framework, in which probes provide the experimenter with an incomplete and biased view of a complex biological system which then allows the construction and validation first of simplifying abstract representations, and consequently of analytical computational frameworks, which themselves provide principles for the design and analysis of biologically-inspired algorithms.

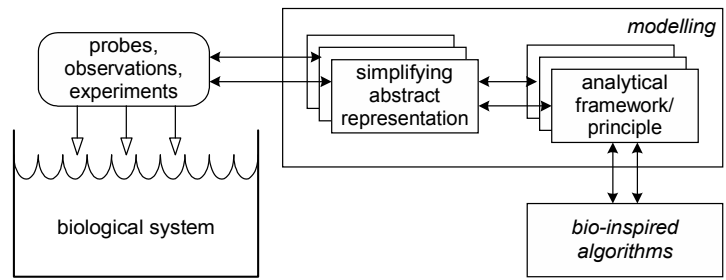

Fig. 4: A conceptual framework for biologically-inspired algorithms [5].

Stepney et al. (ibid.) also apply similar ideas to develop a meta-framework, Figure 5, which allows common underlying properties of classes of models to be analysed by asking questions, called meta-probes, of each of the models under consideration. They suggest a number of questions based around properties which are thought to affect complex behaviour in general. These areas relate to openness, diversity, interaction, structure and scale (ODISS). Using this meta-framework, the authors analyse the commonalities of population and network models.

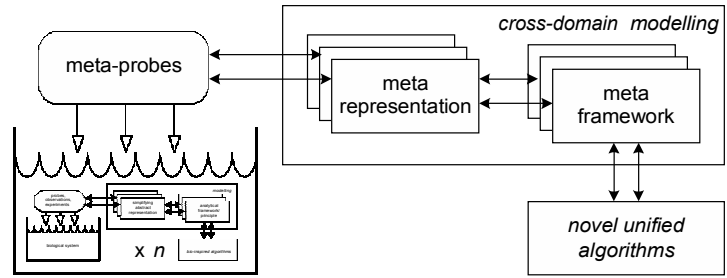

Fig. 5: A conceptual framework for integrating biologically-inspired computational domains [5].

While Stepney et al. use the meta-framework to analyse artificial models for essential features and commonalities, this paper uses it to analyse biological models. The latter approach, apart from being pragmatic as very few artificial models currently exist, also allows biology to have much more of an influence on the meta-framework. Whichever approach is taken, meta-frameworks and the development of computational and mathematical models present a route through which artificial immune system research can help biologists answer research questions in their field.

### 3.2 A meta-framework for innate models

This section takes the general properties of the innate immune system presented in Section 2 and abstracts them by adopting the conceptual framework. The abstracted properties form the basis of a meta-framework for innate models and are presented in terms of each of the ODISS areas of the conceptual framework:

openness : the interaction between the immune system and the host is one of a poised system in dynamic equilibrium coupled to an ever-changing environment. The relatively constant populations of innate immune system cells contrasts with the fluctuating populations of the adaptive system. The innate immune system provides examples of mechanisms for controlling the dynamic allocation of resources of populations of agents.

diversity : the different classes of cells of the innate and adaptive immune systems leads to the idea of distinct groups of functionally similar agents. At a different level, clonal distribution of receptors is an good example of different ways in which diversity manifests itself in biological systems. The underlying processes which drive diversity of innate receptors are evolutionary, while adaptive receptor diversity is established through adaptation.

interaction: in the wider sense considering the innate immune system shows how computation is largely communication, with immunity arising from the cytokine networks of signalling interactions between intercommunicating tissue cells and the innate and adaptive immune systems. Adaptive and innate immune cells are also specialised to access different informational levels. Innate cells focus on class features, while adaptive cells on individual features. Crosstalk between signalling networks is also a prevalent property in the immune system. Spatiality and temporally are key features of interactions across all levels.

structure: considering the innate immune system necessitates a view of the immune system composed of distinct subsystems. Functional similarities as well as differences can be seen between the innate and adaptive subsystems. The innate and adaptive are themselves composed of interacting populations of agents. Cell differentiation pathways provide an even more fine-grained division of cells into types.

scale: diverse populations of large numbers of cells is a hallmark of the immune system. A challenge for artificial immune systems is the need to simulate large populations of agents. Exploiting the emergent properties of distinct populations of large numbers of simple agents rather than a smaller number of more complex agents, along with distributed and parallel architectures for artificial immune systems [28] may provide a way forward.

### 3.3 Refining the framework

As seen in the previous section, the innate immune system provides examples of general properties for artificial systems. Systems of agents form a convenient meta-representation of artificial systems, and many artificial systems are based on populations of interacting agents. This section adopts this meta-representation and refines the general properties of the previous section by discussing how they might be instantiated in models.

Cells seen as autonomous agents forms the basis of the meta-representation discussed here. The intercellular communication involved in all the mechanisms of Section 2 suggests the need for similar means of intercommunication between agents. Signals which allow groups of agents to control the functions and state of other groups of agents are necessary. A finer-grain representation of intercellular signals into distinct classes, as seen in the biological immune system, such as costimulatory, primer or chemokine signals, would allow artificial systems to more closely approximate the control mechanisms and systemic properties of biological systems. A key role of the environment which these agents exist in, termed artificial tissue here, is the provision of a milieu in which agents can interact via signalling. As well as passing signals between agents, mechanisms such as antigen processing and presentation to Th cells by DCs suggest the need for agents with the ability to "consume", process and pass on information to other agents. Some groups of agents, akin to Th or CTL cells, would not have direct access to information, but instead see it through the filter provided by these information processing agents. Artificial tissue would provide the mechanisms for these kinds of interaction with the environment and other agents.

The representation of pathogens at multiple levels suggests another "service" which artificial tissue has to provide. A problem must be represented at multiple levels. The artificial tissue allows agents of the artificial immune system to access different levels of information about events. At the very least, information concerning the structure of events and signals relating to the way elements behave or interact with the tissue as a whole needs to be accessible. Classical static classification problems could perhaps be translated into such a multilevel representation by clustering algorithms or statistical methods which give indications of how individual feature vectors relate to a whole set of other vectors. However, the innate immune system clearly relies on sensing the behaviour as well as structure of pathogens, and tissue models built entirely from information derived from structural considerations, such as similarity or differences between feature vectors, fail to capture this reliance. Dynamic, realtime problems such as intrusion detection offer a much more amenable domain as they naturally include notions of behaviour. For example, a computer virus not only has a particular structure, its program code, but also behaves in a certain way through its interactions with other programs and operating systems, searching for other machines, subverting the function of existing programs, installing backdoors on systems, and so on.

Over its lifetime a cell differentiates along a particular pathway, with each differentiation stage along this pathway representing a specific cell type. All cells at the same stage of differentiation are of the same type and have the same phenotypic configuration and functional characteristics. Which pathway a cell follows is the result of the environmental pressures the cell experiences. Little of the dynamics of the immune system can be captured if agents in artificial immune system models do not possess similar developmental characteristics. This could be modelled by endowing agents with a set of functions, subsets of which the agent performs at any one time and which represent the current type of the agent. Transitions from one type to another are a result of interactions of the agent with its environment and could be pictured as a branching tree structure.

While cells act as individuals, differentiating along their own individual pathways, they also act as part of a group. At this population level, considering the innate immune system highlights the need for groups of agents which respond to different types of information. Certain agents might identify fixed patterns in this information, embodying some type of notional TLR, while others would identify variable patterns, akin to TCRs. The processes which drive the specificity of receptors may be adaptive or evolutionary, with different pressures biasing the type of information surveyed by agents.

Cells control other cells on an individual contact-dependant level. They also control cells in a local neighbourhood through the production of cytokines. This localised control leads to dynamical patterns at the population level. DC control of Th proliferation through costimulatory molecules is a good example of how local interactions control the population of Th cells and determine population-level phenomena such as clonal distribution. Effects of the artificial tissue on one group of agents should have resulting effects on populations of other agents.

The generation of peripheral tolerance by DCs suggests a mechanism by which signals presented by the artificial tissue are received by one group of agents and have a direct effect on other groups of agents. This control might not be as clearcut as live or die, but more a direction of differentiation pathways, of which polarisation of Th cells by TLRs on DCs is a good example.

Lastly, mechanisms of trust or obligation are established. The NK "missing self" model is a good example of this. The provision of sufficient quantities of MHC can be seen as a monitoring requirement, imposed by NK cells, of the system. If tissue cells fail to provide MHC they are destroyed. In realtime monitoring situations, models of such a suppression-based mechanism might be used to establish if groups of data providing agents are functioning.

## 3.4 Summary

Using the biology of the previous section as a basis, this section has sketched out a meta-framework for models of innate immunity, discussing general properties of such models and also how they might be realised more concretely. While the properties presented have tried to capture the core features of innate immunity, due to space and intellectual constraints they are not exhaustive and need to be combined with existing frameworks of adaptive models [5] if integrated models are to be built.

## 4 Conclusion

This paper has presented a summary of current biological understanding of the innate immune system, contrasting it with the adaptive immune system. Adopting a conceptual framework it then proposed and refined a meta-framework for artificial systems incorporating ideas from innate immunity. While emphasising the role of innate immunity, in reality, the innate and adaptive systems are intimately coupled and work together to protect the host. As already suggested, combining the properties suggested here with those of traditional population and network models would enable artificial systems to more closely reflect their biological counterparts.

Other possibilities for future work include a review within the proposed framework of artificial immune system models such as [29, 30] which already include innate immunity. This would help evaluate and compare these models, discerning commonalities and providing direction for future research. Developing more detailed mathematical and computational models would be an important next step in a more detailed understanding of the properties of innate immunity. These models could then be used to instantiate a range of systems in different application domains. More realistic and principled models could also extend understanding on the dynamics of competing immunological models such as those of instructive or consensual regulation of Th1/2 responses, or modulation of costimulatory signals.

Couching ideas of innate immunity within an accepted conceptual framework provides a step in developing more integrated artificial immune system models which take into account the key role the innate immune system plays in host protection. As always, the beauty and subtlety of the immune system will continue to provide a rich source of inspiration for designers of artificial systems.

## 5 Acknowledgements

Many thanks to Adrian Robins for discussions on the immunology, and to my coworkers Uwe Aickelin, Julie Greensmith, Jungwon Kim, Julie McLeod, Steve Cayzer, Rachel Harry, Charlotte Williams, Gianni Tedesco and Peter Bentley, without whom this work would not have been possible. This research is supported by the EPSRC (GR/S47809/01).